\documentclass[10pt,twocolumn,letterpaper]{article}

\usepackage{iccv}
\usepackage{times}
\usepackage{epsfig}
\usepackage{graphicx}
\usepackage{amsmath}
\usepackage{amssymb}

\usepackage{algorithm}
\usepackage{algpseudocode}
\usepackage{amsmath}
\usepackage{graphics}
\usepackage{epsfig}
\usepackage{makecell}
\usepackage{multirow} 

\usepackage[accsupp]{axessibility} 

\usepackage{iccv}
\usepackage{times}
\usepackage{epsfig}
\usepackage{graphicx}
\usepackage{amsmath}
\usepackage{amssymb}

\usepackage{iccv}
\usepackage{times}
\usepackage{epsfig}
\usepackage{graphicx}
\usepackage{amsmath}
\usepackage{amssymb}
\usepackage{graphicx}
\usepackage{amsmath}
\usepackage{amssymb}
\usepackage{booktabs}

\usepackage{algorithm}
\usepackage{algpseudocode}
\usepackage{amsmath}
\usepackage{graphics}
\usepackage{epsfig}
\usepackage{makecell}
\usepackage{multirow} 

\usepackage[breaklinks=true,bookmarks=false]{hyperref}

\iccvfinalcopy 

\def\iccvPaperID{****} 

\ificcvfinal\fi

\begin{document}
\title{ Environment-Invariant Curriculum Relation Learning \\for Fine-Grained Scene Graph Generation}

\author{Yukuan Min\\
	Xidian University\\
	{\tt\small yukuanmin@gmail.com}
\and
Aming Wu$^{*}$\\
Xidian University\\
{\tt\small amwu@xidian.edu.cn}
\and
Cheng Deng\thanks{Corresponding author\\  \indent Codes available at: \textcolor{blue}{https://github.com/myukzzz/EICR}}\\
Xidian University\\
{\tt\small chdeng.xd@gmail.com}
}

\maketitle
\ificcvfinal\thispagestyle{empty}\fi

\begin{abstract}
	The scene graph generation (SGG) task is designed to
	identify the predicates based on the subject-object pairs.  However, existing datasets generally include two imbalance cases: one is the class imbalance from the predicted predicates and another is the context imbalance from the given subject-object pairs, which presents significant challenges for SGG.  Most existing methods focus on the imbalance of the predicted predicate while ignoring the imbalance of the subject-object pairs, which could not achieve satisfactory results. To address the two imbalance cases, we propose a novel Environment Invariant Curriculum Relation learning (EICR) method, which can be applied in a plug-and-play fashion to existing SGG methods. Concretely, to remove the imbalance of the subject-object pairs, we first construct different distribution environments for the subject-object pairs and learn a model invariant to the environment changes. Then, we construct a class-balanced curriculum learning strategy to balance the different environments to remove the predicate imbalance. Comprehensive experiments conducted on VG and GQA datasets demonstrate that our EICR framework can be taken as a general strategy for various SGG models, and achieve significant improvements.
\end{abstract}
\section{Introduction}
\label{sec:intro}

Scene graph generation \cite{graph} (SGG) aims to predict the corresponding predicate (e.g., riding) based on the given subject-object pairs (e.g., (man, bike)). As an intermediate visual understanding task, it can serve as a fundamental tool for high-level vision and language tasks, such as visual question answering \cite{VCTree,VQA,VQA0}, image captioning \cite{caption,caption0,caption1}, and cross-model retrieval \cite{retrival,retrival0}, which promotes the development of visual intelligence.

Though many advances have been achieved \cite{motifs,VCTree}, SGG is still far from satisfactory for practical applications due to the imbalance phenomenon  in the given datasets \cite{TDE}. To this end, most existing methods focus on addressing the class imbalance from the predicted predicates to generate accurate relation words. Particularly, some works propose resampling \cite{DT2-ACBS,BGNN} and reweighting \cite{PCPL} strategies to balance the head and tail predicate classes, which alleviates the imbalance and improves the performance of SGG.

\begin{figure}[t]
	\begin{center}
		\includegraphics[width=1.0\linewidth]{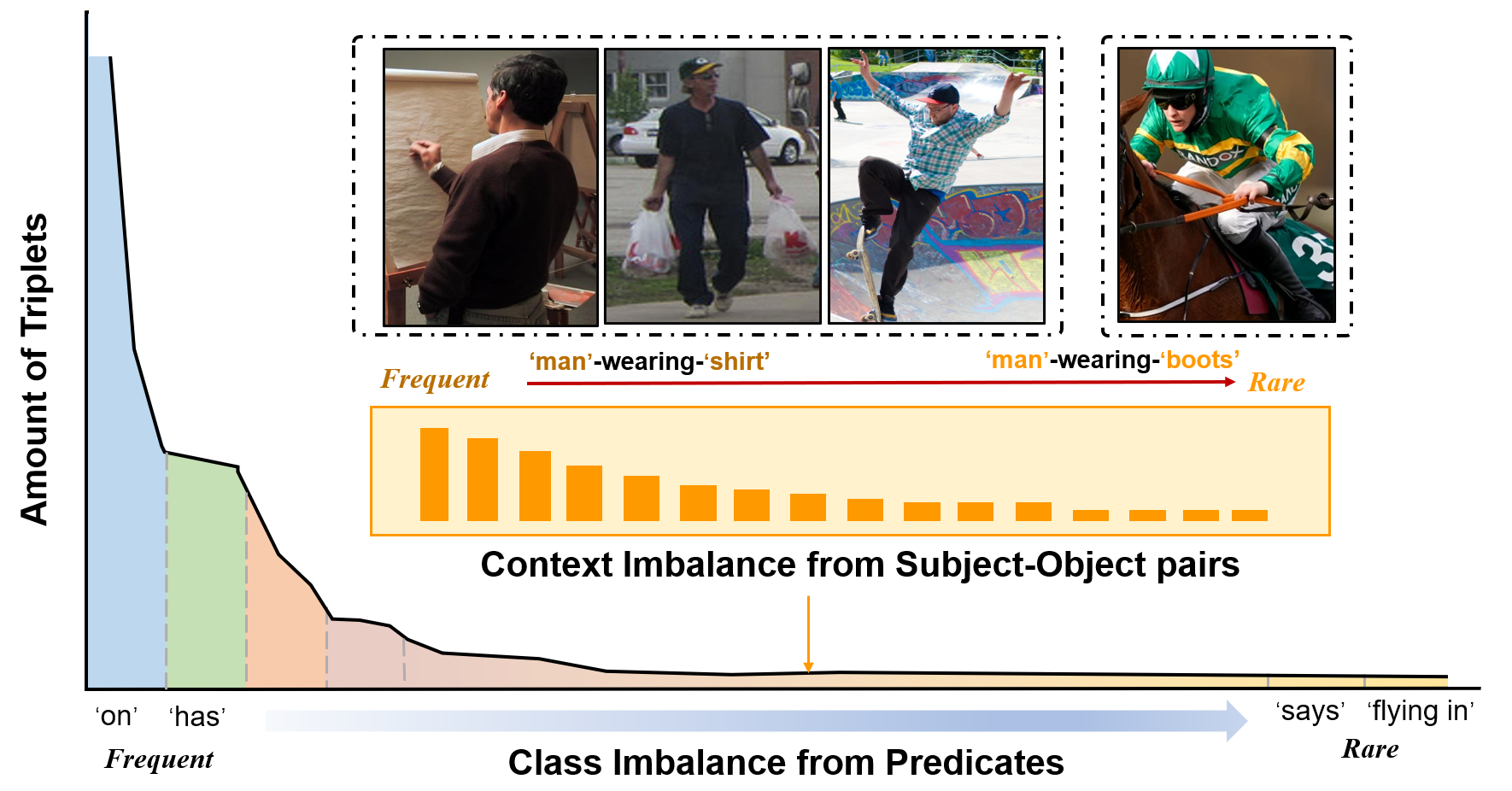}
	\end{center}
	\caption{For SGG datasets, besides the class imbalance from predicates, there exists another imbalance phenomenon, i.e., context imbalance from subject-object pairs, which is easily ignored. To this end, this paper delves into class and context imbalance. And a method of Environment-Invariant Curriculum Relation Learning is proposed to generate fine-grained scene graphs effectively.}
	\label{fig:fig2}
\end{figure}

\begin{figure}[t]
	\begin{center}
		\includegraphics[width=1.0\linewidth]{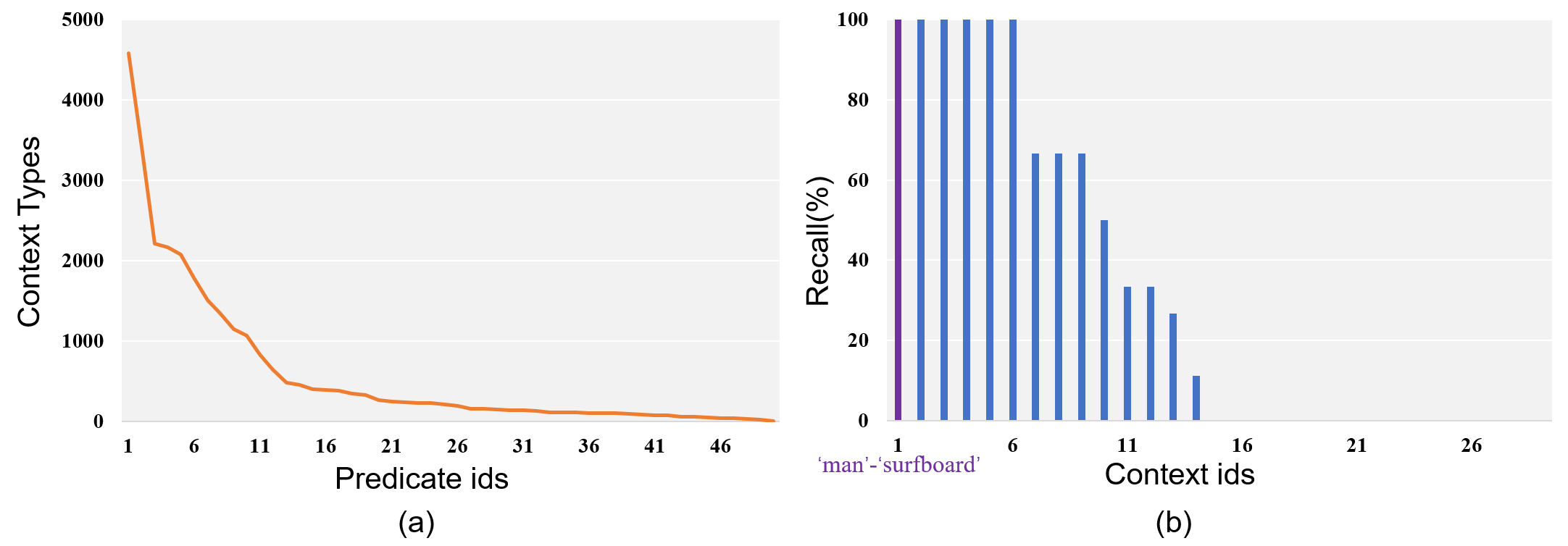}
	\end{center}
	\caption{(a) The number of context subject-object types in all predicate classes.  (b) The Recall@100 performance for different context subject-object in the predicate class `carrying'.}
	\label{fig:fig3}
\end{figure}

Besides the class imbalance from the predicted predicate, there exists another context imbalance from the given subject-object pairs, which is prone to be ignored. As shown in Fig.~\ref{fig:fig2}, since the predicate prediction relies on the given subject-object context, the number imbalance of the given subject-object pairs easily incorrectly predicts the relation between subjects and objects. For example, there exist a large number of relations between `(man, shirt)' and `wearing' in the dataset. When giving a rare subject-object pair, e.g., `(man, boots)', the model is prone to generate an incorrect prediction.   In Fig.~\ref{fig:fig3} (a), we make an analysis of a popular SGG dataset VG \cite{VG}. We observe that the number of context subject-object types will change significantly with the number of predicate categories.  Moreover,  Fig.~\ref{fig:fig3} (b) further shows that performance can be severely affected by the context subject-object. These phenomena show that the current SGG dataset does have the context imbalance problem, and resolving this problem will help produce fine-grained scene graphs and improve the performance.

To address the problems of the two imbalances mentioned above, we propose a novel framework named Environment Invariant Curriculum Relation learning (EICR), which can be equipped with different baseline models in a plug-and-play applied in a fashion.   We construct different distribution environments for the context subject-object and propose a curriculum learning strategy to balance the environments. Specifically, to solve the context imbalance of various subject-object pairs, we construct three different distribution environments: normal, balanced, and over-balanced for the context subject-object pairs, and then apply Invariant Risk Minimization (IRM) \cite{IRM} to learn a context-unbiased relation classifier that is invariant to these environments. To solve the class imbalance, we utilize a class-balanced curriculum learning strategy to first explore the general patterns from the head predicates in the normal environment and then gradually focus on learning the tail predicates in the over-balanced environment.

Our contributions can be summarized as follows:

(1)  Except for the existing class imbalance, we explore and address the under-explored context imbalance problem in the current SGG dataset.

(2) We construct an environment-invariant relation classifier to solve the context imbalance of the subject-object pairs and present a new curriculum learning strategy to consolidate the relation classifier from head to tail predicates and solve the class imbalance of the predicates.

(3) Our EICR can be applied in a plug-and-play fashion for the SGG baselines and get competitive results among various SOTA methods. By applying our proposed method, a VCTree \cite{VCTree} model is improved over \textbf{14\%} on mR@50/100 and over \textbf{12\%} on the metric F@50/100.

\section{Related Work}
\noindent
\textbf{Scene Graph Generation.} SGG provides an efficient way for connecting vision and language \cite{imp,text},  and has drawn widespread attention from the community. Early approaches focus on visual relation detection \cite{VRD,VRD0,VRD1,visualranking} and are mainly dedicated to incorporating more features from various modalities. To further enhance the relations, considering that relations are highly dependent on their context, different methods \cite{messagepassing,motifs,VCTree,contextual} are further proposed to refine the object and relation representations in the scene graph. Motifs \cite{motifs} chose the Bi-LSTM framework for the object and predicate context encoding and VCTree \cite{VCTree} constructs a tree structure to encode the hierarchical and parallel relationships between objects. Moreover, other works also refine the message-passing strategy \cite{messagepassing}. 

\noindent
\textbf{Unbiased Scene Graph Generation.}  Although making steady progress on improving recall on SGG tasks, further research has shown that SGG models are easy to collapse to several general predicate classes because of the long-tail effect in the SGG dataset \cite{biased,unbiased0}. For example, from the causal view \cite{MuLi}, TDE \cite{TDE} employs a causal inference framework to eliminate predicate class bias during the inference process. BGNN \cite{BGNN} constructs a bi-level resampling strategy during the training process.   Inspired by the application of noisy label learning \cite{Yan,Yan1}, NICE \cite{NICE} formulate SGG as an out-of-distribution detection \cite{MuLi1,MuLi2} problem and propose a noisy label correction strategy for unbiased SGG. Different from existing SGG works,  we are the first to explicitly define and address the context imbalance of the subject-object pairs on SGG datasets.

\section{EICR for Class and Context Imbalances}

For SGG, this paper aims to address the two different kinds of distribution imbalance, i.e., class imbalance of predicates and  context imbalance of subject-object pairs.

\begin{figure*}[t]
	\begin{center}
		\includegraphics[width=0.9\linewidth]{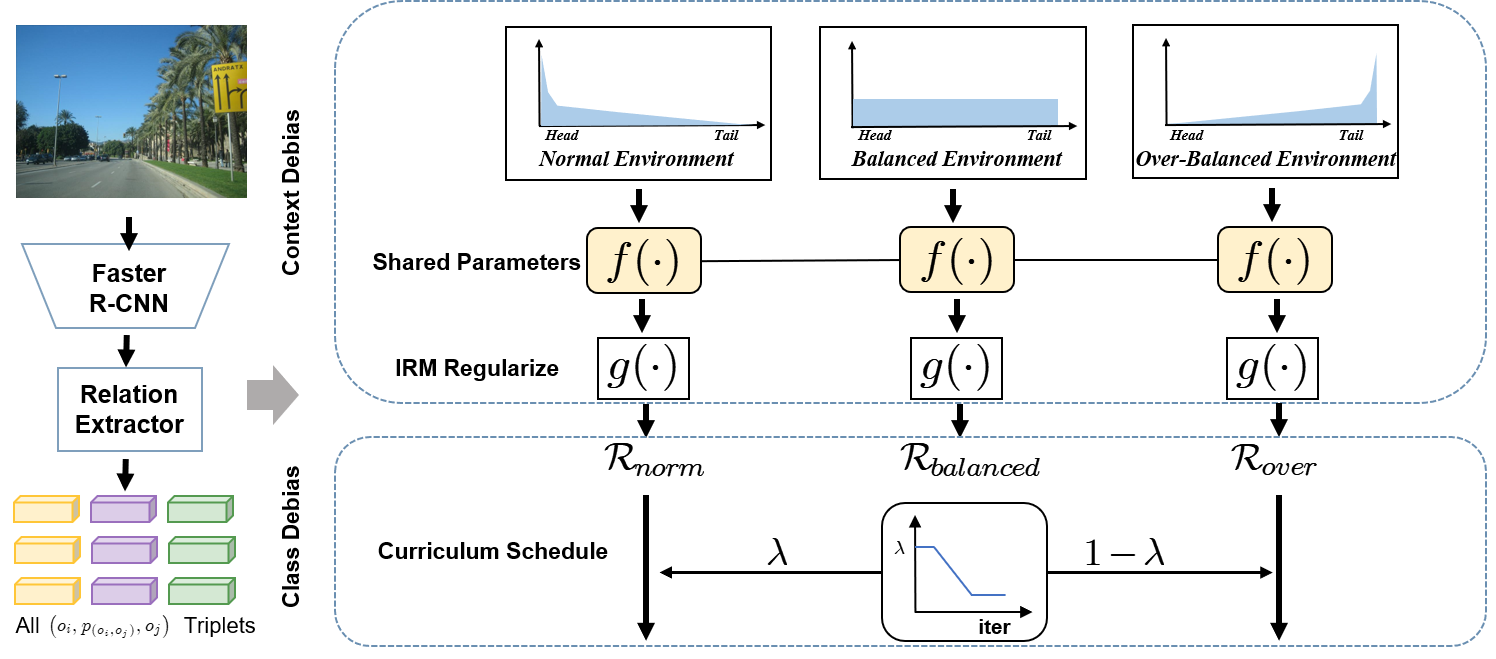}
	\end{center}
	\caption{Illustration of our method for alleviating the context and class imbalances in SGG. Firstly, an environment-invariant learning module to build multiple different distribution environments of subject-object pairs, which is beneficial for obtaining an invariant relation classifier and alleviating the context imbalance. Next, a class-balanced curriculum learning strategy is designed to balance the built multiple environments, alleviating the class imbalance.}
	\label{fig:fig1}
\end{figure*}

\subsection{Preliminary}

The scene graph generation task aims to generate a summary graph $\mathcal{G}$ for the given image $I$. Specifically,  a scene graph $\mathcal{G}=\left\{\left(O, E\right)\right\}$ corresponding to $I$ contains a set of target entities $O=\left\{\left(o_i\right)\right\}_{i=1}^{N_o}$ and a set of relational triplets $E=\left\{\left(o_i, p_{\left(o_i, o_j\right)}, o_j\right)\right\}_{i,j=1}^{N_e}$, where $o_i \in O$ and $o_j \in O$, $p_{\left(o_i, o_j\right)}$ is defined as the relation between them and belongs to the predefined predicate class set $\mathcal{P}$.

Classifying a relation $p$ as the predicate class $c$ can be preliminarily defined as predicting $P(r=c\mid p)$ based on the dataset of the relations and its label pairs $\{(p, r)\}$ \cite{motifs}. Using Bayes theorem \cite{bayesian}, the predictive model could be decomposed as $P(r=c\mid p)=\frac{P(p\mid r=c) \cdot P(r=c)}{P(p)}$, where $P(r=c)$ is the class distribution, and $P(p)$ is the marginal distribution of the relations. Previous SGG methods \cite{TDE,BGNN}  only consider class imbalance $P(r=c)$ while ignoring the context imbalance from the different marginal distribution $P(p)$ based on subject-object pairs. 

To explicitly define the relation classification $P(r=c\mid p)$, we assume that a relation $p$ is generated by a set of hidden attributes $z=\left\{z_1, z_2, z_3, \ldots\right\}$. Since there exists predicate class imbalance and context imbalance in the existing SGG dataset, we defined two disjoint subsets for the hidden attributes: class-specific attributes $z_c$ (e.g., the frequency of the predicate classes \cite{TDE}) and context-specific attributes $z_e$ (e.g., the various context subject-object pairs $\{o_i,o_j\}$).  Thus we can further decompose the relation prediction model $P\left(r=c \mid z_c, z_e\right)$ as follows:
\begin{equation}
	\begin{aligned}
		P\left(r=c \mid z_c, z_e\right)=\frac{P\left(z_c \mid r=c\right)}{P\left(z_c,z_e\right)}\\
		\cdot \underbrace{P\left(z_e \mid r=c, z_c\right)}_{\text {context imbalance }} \cdot \underbrace{P(r=c)}_{\text {class imbalance}}.
	\end{aligned}
	\label{eq1}
\end{equation}
From Eq. \ref{eq1}, the relation classifier $P\left(r=c \mid z_c, z_e\right)$ is affected by two imbalances: 

\textbf{Class Imbalance}: in previous SGG works \cite{BGNN,TDE,PCPL}, the distribution of $P(r=c)$ is considered as the main cause of the performance degradation. As $P(r=c)$ can be explicitly calculated from the training data, the majority of previous methods directly alleviate its effect by class-wise adjustment \cite{PCPL} or re-sampling \cite{BGNN,SHA}. 

\textbf{Context Imbalance}: we argue that the context imbalance suffers the relation classifier in two ways. First, as shown in Fig.~\ref{fig:fig3} (a), the different diversity for the context makes imbalanced influences for the predicate classes. Second, Fig.~\ref{fig:fig3} (b) shows that the predicate is high-related to certain context subject-object pairs. These phenomenons will create spurious correlations between the subject-object pairs and the predicates, which will weaken the relation classification performance, especially in the tail predicates whose context subject-object pairs are rare due to sampling scarcity. Since we have concluded that the context subject-object pair is highly related to the predicate class,  we formulate the context imbalance as $P\left(z_e \mid r=c, z_c\right)$.

To solve these two imbalances and obtain unbiased relations for the scene graphs, we propose the following Environment Invariant Curriculum Relation learning (EICR) framework to learn an unbiased relation classifier invariant to the change of the various predicate classes and context subject-object pairs.

\subsection{Environment Invariant Learning}
To alleviate the context imbalance, we need to eliminate the impact of the contexts for the relation classification $P\left(z_e \mid r=c, z_c\right)$. To this end,  based on the theory of Invariant Risk Minimization (IRM) \cite{IRM},  we can first construct a set of environments $\mathcal{E}=\left\{e_1, e_2, \ldots\right\}$ with diverse context distribution. Then, by regularizing the relation classifier $g(\cdot)$ to be equally optimal across the environments with different context distributions, we can alleviate the influence of the context. Thus objective function can be defined as follows:
\begin{equation}
	\begin{aligned}
		&\min _g \sum_{e \in \mathcal{E}} R^e(I, r ; f(\cdot), g(\cdot)),\\
		&\text { subject to } g \in \arg \min _g R^e \text { for all } e \in \mathcal{E},
	\end{aligned}
	\label{eq2}
\end{equation}
where  $R^e(I, r ; f(\cdot), g(\cdot))$ is the risk under environment $e$ (i.e., the loss for training), $f(\cdot)$ is the relation feature extractor, $g \in \arg \min _g R^e(I, r ; f(\cdot), g(\cdot))$ for all $e \in \mathcal{E}$ means that the invariant identifier $g$ should minimize the risk under all environments simultaneously. Following IRM \cite{IRM}, we use a gradient norm penalty term to minimize $g$ at each environment, i.e., $\min _{\Phi: \mathcal{X} \rightarrow \mathcal{Y}} \sum_{e \in \mathcal{E}} R^e(\Phi)+ \lambda \cdot\left\|\nabla_{w \mid w=1.0} R^e(w \cdot \Phi)\right\|^2$,where $\Phi$ is the invariant model, $R(\cdot)$ denotes the training loss under different environments $e \in \mathcal{E}$ and we set $\lambda=1$.  The detailed process of the environment construction is introduced below:

The set of diverse environments should ensure the variance of the context influence and ideally are orthogonal distributions \cite{NIPS21hwzhang,yfZhang}. However, considering the computation consumption and feasibility of the strategy, to construct different $P\left(z_e \mid r=c, z_c\right)$, it is hard to change the $r=c$ since the relation labels are predefined. Thus we construct three learning environments with different $z_c$, i.e., the frequency of the predicate classes. As illustrated in Fig.~\ref{fig:fig1}, each learning environment constructs different frequencies of the predicate classes: 

$\bullet$ The normal environment maintains the raw distribution of the predicate classes in the dataset. 

$\bullet$ The class-balanced environment constructs the resampling strategy \cite{SHA}  to sampling in all predicate categories at balanced frequencies. Specifically,  we first calculate the median amount of the samples over all predicate classes $Median(r)$. Then, for each predicate class $r_i$ with $n_i$ samples, we calculate the sampling rate $s_i$ as follows:
\begin{equation}
	s_i= \begin{cases}\frac{{Median}\left(r\right)}{n_i} & \text { if } Median(r) \leq n_i, \\ 1 & \text { if } Median(r) > n_i. \end{cases}
\end{equation}

$\bullet$ The over-balanced environment is constructed to over-correct the imbalanced predicate class distribution $P(r)$. Thus we first construct a resampling strategy for the balanced sampling as in the class-balanced environment. Then we adopt an extra reweighting strategy for over-balanced weighting, the loss can be formulated as:
\begin{equation}
	L_{over}=-\sum_{i=1}^C w_i r_i\log(g(f(I))),
\end{equation}
where $w_i=1/n_i$ and $C$ denotes the number of the predicate categories. This environment deliberately picks relation triplets with the probability negatively correlated with predicate class size. 

\begin{algorithm}[t]
	\caption{EICR Framework} 
	\hspace*{0.02in} {\bf Input:} 
	SGG Dataset $\{(I,r)\}$,  \# Iteration $T$.
	\begin{algorithmic}[1]
		\State Initialize the pretrained relation feature extractor $f$ and relation classifier $g$
		
		\While{$t \leq T$}
		
		\State // \emph{Context-Debias}
		\State Generate multiple environments $\left\{e_1, e_2, e_3 \right\}$  
		\State Learn parameters of $g$ through IRM with Eq. ~\ref{eq2}
		\State // \emph{Class-Debias}
		\State Reweight environments by schedule  in Eq. ~\ref{eq:eq3} 
		\State Update the model by balanced risks from Eq. ~\ref{eq:eq5}.

		\EndWhile

	\end{algorithmic}
	\hspace*{0.02in} {\bf Output:} 
	The debiased relation feature extractor $f$ and relation classifier $g$
	\label{alg:1}

\end{algorithm}
\begin{table*}
	\centering
	\footnotesize 
	\setlength{\tabcolsep}{1mm}{
		\begin{tabular}{cc|ccc|ccc|ccc}
			\hline
			\multicolumn{2}{c|}{
				\multirow{2}*{\textbf{Method}}}  & \multicolumn{3}{c|}{\makecell[c]{\textbf{PredCls}}} & \multicolumn{3}{c|}{\textbf{SGCls}} & \multicolumn{3}{c}{\textbf{SGDet}}  \\
			\Xcline{3-11}{0.5pt}
			& &\makecell[c]R@50 / 100 &\makecell[c]mR@50 / 100& \makecell[c]F@50 / 100 & \makecell[c]R@50 / 100 &\makecell[c]mR@50 / 100& \makecell[c]F@50 / 100
			& \makecell[c]R@50 / 100 &\makecell[c]mR@50 / 100& \makecell[c]F@50 / 100 \\
			
			\hline
			
			\multicolumn{2}{c|}{\makecell[l]{IMP \cite{imp}}} & \makecell[c]{61.1 / 63.1} & \makecell[c]{11.0 / 11.8} & \makecell[c]{18.6 / 19.9} & \makecell[c]{37.4 / 38.3}&\makecell[c]{6.4 / 6.7}&\makecell[c]{10.9 / 11.4}&\makecell[c]{23.6 / 28.7}&\makecell[c]{3.3 / 4.1}& \makecell[c]{5.8 / 7.2}\\
			
			\multicolumn{2}{c|}{\makecell[l]{GPS-Net \cite{GPSNet}}} & \makecell[c]{65.2 / 67.1} & \makecell[c]{15.2 / 16.6} & \makecell[c]{24.7 / 26.6} & \makecell[c]{37.8 / 39.2}&\makecell[c]{8.5 / 9.1}&\makecell[c]{13.9 / 14.8}&\makecell[c]{31.1 / 35.9}&\makecell[c]{6.7 / 8.6}& \makecell[c]{18.9 / 22.3}\\

			\multicolumn{2}{c|}{\makecell[l]{BGNN \cite{BGNN}}} & \makecell[c]{59.2 / 61.3} & \makecell[c]{30.4 / 32.9} & \makecell[c]{40.2 / 42.8} & \makecell[c]{37.4 / 38.5}&\makecell[c]{14.3 / 16.5}&\makecell[c]{20.7 / 23.1} &\makecell[c]{31.0 / 35.8} &\makecell[c]{10.7 / 12.6} &\makecell[c]{15.9 / 18.6} \\
			
			\multicolumn{2}{c|}{\makecell[l]{DT2-ACBS \cite{DT2-ACBS}}} & \makecell[c]{23.3 / 25.6} & \makecell[c]{35.9 / 39.7} & \makecell[c]{28.3 / 31.1} & \makecell[c]{16.2 / 17.6}&\makecell[c]{24.8 / 27.5}&\makecell[c]{19.6 / 21.5} &\makecell[c]{15.0 / 16.3} &\makecell[c]{22.0 / 24.0} &\makecell[c]{17.8 / 19.4} \\
			
			\multicolumn{2}{c|}{\makecell[l]{SHA-GCL \cite{SHA}}} & \makecell[c]{35.1 / 37.2} & \makecell[c]{41.6 / 44.1} & \makecell[c]{38.1 / 40.4} & \makecell[c]{22.8 / 23.9}&\makecell[c]{23.0 / 24.3}&\makecell[c]{22.9 / 24.1} &\makecell[c]{14.9 / 18.2} &\makecell[c]{17.9 / 20.9} &\makecell[c]{16.3 / 19.5} \\

			\hline
			
			\multicolumn{2}{c|}{\makecell[l]{Motifs \cite{motifs}}} & \makecell[c]{65.2 / 67.0} & \makecell[c]{14.8 / 16.1} & \makecell[c]{24.1 / 26.0} & \makecell[c]{38.9 / 39.8}&\makecell[c]{8.3 / 8.8}&\makecell[c]{13.7 / 14.8} &\makecell[c]{31.1 / 35.9} &\makecell[c]{6.7 / 8.6} &\makecell[c]{11.0 / 13.9} \\

			\multicolumn{2}{c|}{\makecell[l]{ + TDE \cite{TDE}}} & \makecell[c]{46.2 / 51.4} & \makecell[c]{25.5 / 29.1} & \makecell[c]{32.9 / 37.2} & \makecell[c]{27.7 / 29.9}&\makecell[c]{13.1 / 14.9}&\makecell[c]{17.8 / 19.9} &\makecell[c]{16.9 / 20.3} &\makecell[c]{8.2 / 9.8} &\makecell[c]{11.0 / 13.2} \\		
			
			\multicolumn{2}{c|}{\makecell[l]{ + PCPL \cite{PCPL}}} & \makecell[c]{54.7 / 56.5} & \makecell[c]{24.3 / 26.1} & \makecell[c]{33.7 / 35.7} & \makecell[c]{35.3 / 36.1}&\makecell[c]{12.0 / 12.7}&\makecell[c]{17.9 / 18.8} &\makecell[c]{27.8 / 31.7} &\makecell[c]{10.7 / 12.6} &\makecell[c]{15.5 / 18.0} \\

			\multicolumn{2}{c|}{\makecell[l]{ + EBM \cite{EBM}}} & \makecell[c]{65.2 / 67.3} & \makecell[c]{18.0 / 19.5} & \makecell[c]{28.2 / 30.2} & \makecell[c]{39.2 / 40.0}&\makecell[c]{10.2 / 11.0}&\makecell[c]{16.2 / 17.3} &\makecell[c]{31.7 / 36.3} &\makecell[c]{7.7 / 9.3} &\makecell[c]{12.4 / 14.8} \\

			\multicolumn{2}{c|}{\makecell[l]{ + NICE \cite{NICE}}} & \makecell[c]{55.1 / 57.1} & \makecell[c]{29.9 / 32.3} & \makecell[c]{38.8 / 41.3} & \makecell[c]{33.1 / 34.0}&\makecell[c]{16.6 / 17.9}&\makecell[c]{22.1 / 23.5} &\makecell[c]{27.8 / 31.8} &\makecell[c]{12.2 / 14.4} &\makecell[c]{17.0 / 19.8} \\

			\multicolumn{2}{c|}{\makecell[l]{ + IETrans \cite{IETrans}}} & \makecell[c]{48.6 / 50.5} & \makecell[c]{35.8 / 39.1} & \makecell[c]{41.2 / 44.1} & \makecell[c]{29.4 / 30.2}&\makecell[c]{21.5 / 22.8}&\makecell[c]{24.8 / 26.0} &\makecell[c]{23.5 / 27.2} &\makecell[c]{15.5 / 18.0} &\makecell[c]{18.7 / 21.7} \\	
			
			\multicolumn{2}{c|}{\makecell[l]{ + EICR}} & \makecell[c]{55.3 / 57.4} & \makecell[c]{34.9 / 37.0} & \makecell[c]{\textbf{42.8 / 45.0}} & \makecell[c]{34.5 / 35.4}&\makecell[c]{20.8 / 21.8}&\makecell[c]{\textbf{25.9 / 27.0}} &\makecell[c]{27.9 / 32.2} &\makecell[c]{\textbf{15.5 / 18.2}} &\makecell[c]{\textbf{19.9 / 23.3}} \\			
			
			\hline
			
			\multicolumn{2}{c|}{\makecell[l]{VCTree \cite{VCTree}}} & \makecell[c]{65.4 / 67.2} & \makecell[c]{16.7 / 18.2} & \makecell[c]{26.6 / 28.6} & \makecell[c]{46.7 / 47.6}&\makecell[c]{11.8 / 12.5}&\makecell[c]{18.8 / 19.8} &\makecell[c]{31.9 / 36.2} &\makecell[c]{7.4 / 8.7} &\makecell[c]{12.0 / 14.0} \\

			\multicolumn{2}{c|}{\makecell[l]{ + TDE \cite{TDE}}} & \makecell[c]{47.2 / 51.6} & \makecell[c]{25.4 / 28.7} & \makecell[c]{33.0 / 36.9} & \makecell[c]{25.4 / 27.9}&\makecell[c]{12.2 / 14.0}&\makecell[c]{16.5 / 18.6} &\makecell[c]{19.4 / 23.2} &\makecell[c]{9.3 / 11.1} &\makecell[c]{12.6 / 15.1} \\		
			
			\multicolumn{2}{c|}{\makecell[l]{ + PCPL \cite{PCPL}}} & \makecell[c]{56.9 / 58.7} & \makecell[c]{22.8 / 24.5} & \makecell[c]{32.6 / 34.6} & \makecell[c]{40.6 / 41.7}&\makecell[c]{15.2 / 16.1}&\makecell[c]{22.1 / 23.2} &\makecell[c]{19.4 / 23.2} &\makecell[c]{9.3 / 11.1} &\makecell[c]{12.6 / 15.0} \\

			\multicolumn{2}{c|}{\makecell[l]{ + EBM \cite{EBM}}} & \makecell[c]{64.0 / 65.8} & \makecell[c]{18.2 / 19.7} & \makecell[c]{28.3 / 30.3} & \makecell[c]{44.7 / 45.8}&\makecell[c]{12.5 / 13.5}&\makecell[c]{19.5 / 20.9} &\makecell[c]{31.4 / 35.9} &\makecell[c]{7.7 / 9.1} &\makecell[c]{12.4 / 14.5} \\

			\multicolumn{2}{c|}{\makecell[l]{ + NICE \cite{NICE}}} & \makecell[c]{55.0 / 56.9} & \makecell[c]{30.7 / 33.0} & \makecell[c]{39.4 / 41.8} & \makecell[c]{37.8 / 39.0}&\makecell[c]{19.9 / 21.3}&\makecell[c]{26.1 / 27.6} &\makecell[c]{27.0 / 30.8} &\makecell[c]{11.9 / 14.1} &\makecell[c]{16.5 / 19.3} \\	
			
			\multicolumn{2}{c|}{\makecell[l]{ + IETrans \cite{IETrans}}} & \makecell[c]{48.0 / 49.9} & \makecell[c]{37.0 / 39.7} & \makecell[c]{41.8 / 44.2} & \makecell[c]{30.0 / 30.9}&\makecell[c]{19.9 / 21.8}&\makecell[c]{23.9 / 25.6} &\makecell[c]{23.6 / 27.8} &\makecell[c]{12.0 / 14.9} &\makecell[c]{15.9 / 19.4} \\

			\multicolumn{2}{c|}{\makecell[l]{ + EICR}} & \makecell[c]{56.0 / 57.9} & \makecell[c]{35.6 / 37.9} & \makecell[c]{\textbf{43.6/ 45.8}} & \makecell[c]{39.4 / 40.5}&\makecell[c]{\textbf{26.2 / 27.4}}&\makecell[c]{\textbf{32.8 / 33.9}} &\makecell[c]{26.0 / 30.1} &\makecell[c]{\textbf{15.2 / 17.5}} &\makecell[c]{\textbf{19.2 / 22.1}} \\

			\hline

			\multicolumn{2}{c|}{\makecell[l]{Transformer \cite{TDE}}} & \makecell[c]{63.6 / 65.7} & \makecell[c]{19.7 / 19.6} & \makecell[c]{27.9 / 30.2} & \makecell[c]{38.1 / 39.2}&\makecell[c]{9.9 / 10.5}&\makecell[c]{15.7 / 16.6} &\makecell[c]{30.0 / 34.3} &\makecell[c]{7.4 / 8.8} &\makecell[c]{11.9 / 14.0} \\	
			
			\multicolumn{2}{c|}{\makecell[l]{ + CogTree \cite{cogtree}}} & \makecell[c]{38.4 / 39.7} & \makecell[c]{28.4 / 31.0} & \makecell[c]{32.7 / 34.8} & \makecell[c]{22.9 / 23.4}&\makecell[c]{15.7 / 16.7}&\makecell[c]{18.6 / 19.5} &\makecell[c]{19.5 / 21.7} &\makecell[c]{11.1 / 12.7} &\makecell[c]{14.1 / 16.0} \\

			\multicolumn{2}{c|}{\makecell[l]{ + IETrans \cite{IETrans}}} & \makecell[c]{49.0 / 50.8} & \makecell[c]{35.0 / 38.0} & \makecell[c]{40.8/ 43.5} & \makecell[c]{29.6 / 30.5}&\makecell[c]{20.8 / 22.3}&\makecell[c]{24.4 / 25.8} &\makecell[c]{23.1 / 27.1} &\makecell[c]{15.0 / 18.1} &\makecell[c]{18.2 / 21.7} \\	
			
			\multicolumn{2}{c|}{\makecell[l]{ + EICR}} & \makecell[c]{52.8 / 54.7} & \makecell[c]{\textbf{36.9 / 39.1}} & \makecell[c]{\textbf{43.5/ 45.6}} & \makecell[c]{31.4 / 32.4}&\makecell[c]{\textbf{21.6 / 22.4}}&\makecell[c]{\textbf{25.6 / 26.5}} &\makecell[c]{23.7 / 27.7} &\makecell[c]{\textbf{17.3 / 19.7}} &\makecell[c]{\textbf{20.0 / 23.0}} \\			
			
			\hline

			\\
			
	\end{tabular}}
	
	\caption{Performance (\%) of our method and other baselines on VG dataset. + EICR denotes different models equipped with our EICR. }
	\label{ta:1}
\end{table*}
\begin{table*}
	\centering
	\footnotesize 
	\setlength{\tabcolsep}{1mm}{
		\begin{tabular}{cc|ccc|ccc|ccc}
			\hline
			\multicolumn{2}{c|}{
				\multirow{2}*{\textbf{Method}}}  & \multicolumn{3}{c|}{\makecell[c]{\textbf{PredCls}}} & \multicolumn{3}{c|}{\textbf{SGCls}} & \multicolumn{3}{c}{\textbf{SGDet}}  \\
			\Xcline{3-11}{0.5pt}
			& &\makecell[c]R@50 / 100 &\makecell[c]mR@50 / 100& \makecell[c]F@50 / 100 & \makecell[c]R@50 / 100 &\makecell[c]mR@50 / 100& \makecell[c]F@50 / 100
			& \makecell[c]R@50 / 100 &\makecell[c]mR@50 / 100& \makecell[c]F@50 / 100 \\
			
			\hline
			
			\multicolumn{2}{c|}{\makecell[l]{Motifs \cite{motifs}}} & \makecell[c]{65.3 / 66.8} & \makecell[c]{16.4 / 17.1} & \makecell[c]{26.2 / 27.2} & \makecell[c]{34.2 / 34.9}&\makecell[c]{8.2 / 8.6}&\makecell[c]{13.2 / 13.8}&\makecell[c]{28.9 / 33.1}&\makecell[c]{6.4 / 7.7}& \makecell[c]{10.5 / 12.5}\\
			
			\multicolumn{2}{c|}{\makecell[l]{ + GCL \cite{SHA}}} & \makecell[c]{44.5 / 46.2} & \makecell[c]{36.7 / 38.1} & \makecell[c]{40.2 / 41.8} & \makecell[c]{23.2 / 24.0}&\makecell[c]{17.3 / 18.1}&\makecell[c]{19.8 / 20.6}&\makecell[c]{18.5 / 21.8}&\makecell[c]{16.8 / 18.8}& \makecell[c]{17.6 / 20.2}\\
			
			\multicolumn{2}{c|}{\makecell[l]{ + EICR}} & \makecell[c]{56.4 / 58.1} & \makecell[c]{36.3 / 38.0} & \makecell[c]{\textbf{44.2 / 46.0}} & \makecell[c]{28.8 / 29.4}&\makecell[c]{17.2 / \textbf{18.2}}&\makecell[c]{\textbf{21.5 / 22.5}}&\makecell[c]{24.6 / 28.4}&\makecell[c]{16.0 / 18.0}& \makecell[c]{\textbf{19.4 / 22.0}}\\

			\hline
			
			\multicolumn{2}{c|}{\makecell[l]{VCTree \cite{VCTree}}} & \makecell[c]{63.8 / 65.7} & \makecell[c]{16.6 / 17.4} & \makecell[c]{26.3 / 27.5} & \makecell[c]{34.1 / 34.8}&\makecell[c]{7.9 / 8.3}&\makecell[c]{12.8 / 13.4}&\makecell[c]{28.3 / 31.9}&\makecell[c]{6.5 / 7.4}& \makecell[c]{10.6 / 13.2}\\
			
			\multicolumn{2}{c|}{\makecell[l]{ + GCL \cite{SHA}}} & \makecell[c]{44.8 / 46.6} & \makecell[c]{35.4 / 36.7} & \makecell[c]{39.5 / 41.1} & \makecell[c]{23.7 / 24.5}&\makecell[c]{17.3 / 18.0}&\makecell[c]{20.0 / 20.8}&\makecell[c]{17.6 / 20.7}&\makecell[c]{15.6 / 17.8}& \makecell[c]{16.5 / 19.1}\\
			
			\multicolumn{2}{c|}{\makecell[l]{ + EICR}} & \makecell[c]{55.3 / 57.0} & \makecell[c]{\textbf{35.9 / 37.4}} & \makecell[c]{\textbf{43.5 / 45.2}} & \makecell[c]{28.4 / 29.1}&\makecell[c]{\textbf{17.8 / 18.6}}&\makecell[c]{\textbf{21.9 / 22.7}}&\makecell[c]{24.0 / 27.6}&\makecell[c]{14.7 / 16.3}& \makecell[c]{\textbf{18.2 / 20.5}}
			
			\\

			\hline

			\\
			
	\end{tabular}}
	
	\caption{Performance comparison of different methods on three tasks of GQA dataset}
	\label{ta:2}
\end{table*}

\subsection{Class-Balanced Curriculum Learning}
After obtaining a context-unbiased relation representation from the environment learning module, we assume the network has already modeled the  $P\left(z_e \mid r=c, z_c\right)$. Therefore, we only need to tackle the class imbalance $P(r=c)$ in the context-balanced SGG data. We devise a curriculum schedule for environment learning to make the relation prediction model successfully explore general patterns from head predicates and then gradually focus on the tail predicates. Specifically, we adjust the learning weights between the over-balanced environment and the normal environment by a trade-off factor $\lambda$ which is defined as:
\begin{equation}
	\lambda= \begin{cases}\lambda_{max} & \text { if } t \leq T, \\ \max \left(H(t), \lambda_{min}\right) & \text { if } T<t \leq 2T,  \\\lambda_{min} & \text { if } t>2T,\end{cases}
	\label{eq:eq3}
\end{equation}
where $t$ is the current training iteration, $T$ is predefined intermediate training iterations for different stages of curriculum learning. $\lambda_{min},\lambda_{max} \in[0,1]$ are hyper-parameters. In order to ensure the scale invariance, $\lambda_{min}+\lambda_{max}=1$. $H(t)$ is  a curriculum schedule function decreasing from 1 to 0 with the input iteration $t$, which can be defined as:
\begin{equation}
	H(t)=\frac{2T-t}{T}(\lambda_{max}-\lambda_{min}),
	\label{eq:eq4}
\end{equation}
thus the joint loss function for the three environments can be formulated as:
\begin{equation}
	\mathcal{R}_{\text {hybird}}=\lambda \cdot \mathcal{R}_{norm}+(1-\lambda) \cdot \mathcal{R}_{over}+ \mathcal{R}_{balanced},
	\label{eq:eq5}
\end{equation}
where $\mathcal{R}_{norm},\mathcal{R}_{over},\mathcal{R}_{balanced}$ are the risks under normal, class-balanced, and over-balanced environments. The class-balanced curriculum learning strategy thus can be divided into three phases.  In the first training phase ($t \leq T$), the model is mainly focused on the normal environment to learn the general patterns from head predicates. In the second phase ($T<t \leq 2T $), $\lambda$ gradually decreases during the training. The model’s learning focus shifts from the normal environment to the over-balanced environment to incrementally learn the fine-grained tail predicates while retaining the general patterns.  In the third phase ($t>2T$), the model avoids overfitting the general patterns from the normal environment when focusing on the tail predicates at later training periods. Algorithm. ~\ref{alg:1} shows details of EICR.


\section{Experiments}

In this section, we first show the generalizability of our method with different baseline models and the expansibility to different SGG datasets. Ablation studies are also constructed to explore the influence of different modules and hyperparameters. Finally, we conduct several analyses to show the effectiveness of our method in solving both the context imbalance and the class imbalance.

\subsection{Experimental Settings}
\begin{table*}
	\centering
	\footnotesize 
	\setlength{\tabcolsep}{1mm}{
		\begin{tabular}{cc|ccc|ccc|ccc}
			\hline
			\multicolumn{2}{c|}{
				\multirow{2}*{\textbf{Method}}}  & \multicolumn{3}{c|}{\makecell[c]{\textbf{PredCls}}} & \multicolumn{3}{c|}{\textbf{SGCls}} & \multicolumn{3}{c}{\textbf{SGDet}}  \\
			\Xcline{3-11}{0.5pt}
			& &\makecell[c]R@50 / 100 &\makecell[c]mR@50 / 100& \makecell[c]F@50 / 100 & \makecell[c]R@50 / 100 &\makecell[c]mR@50 / 100& \makecell[c]F@50 / 100
			& \makecell[c]R@50 / 100 &\makecell[c]mR@50 / 100& \makecell[c]F@50 / 100 \\

			\hline
			
			\multicolumn{2}{c|}{\makecell[l]{Motifs \cite{motifs}}} & \makecell[c]{65.2 / 67.0} & \makecell[c]{14.8 / 16.1} & \makecell[c]{24.1 / 26.0} & \makecell[c]{38.9 / 39.8}&\makecell[c]{8.3 / 8.8}&\makecell[c]{13.7 / 14.8} &\makecell[c]{31.1 / 35.9} &\makecell[c]{6.7 / 8.6} &\makecell[c]{11.0 / 13.9} \\

			\multicolumn{2}{c|}{\makecell[l]{ + TDE \cite{TDE}}} & \makecell[c]{46.2 / 51.4} & \makecell[c]{25.5 / 29.1} & \makecell[c]{32.9 / 37.2} & \makecell[c]{27.7 / 29.9}&\makecell[c]{13.1 / 14.9}&\makecell[c]{17.8 / 19.9} &\makecell[c]{16.9 / 20.3} &\makecell[c]{8.2 / 9.8} &\makecell[c]{11.0 / 13.2} \\

			\multicolumn{2}{c|}{\makecell[l]{ + EIL }} & \makecell[c]{64.1 / 65.8} & \makecell[c]{24.5 / 26.5} & \makecell[c]{\textbf{35.5 / 37.8}} & \makecell[c]{39.3 / 40.1}&\makecell[c]{\textbf{15.4 / 16.1}}&\makecell[c]{\textbf{22.1 / 23.0}} &\makecell[c]{32.2 / 36.8} &\makecell[c]{\textbf{10.6 / 12.6}} &\makecell[c]{\textbf{15.9 / 18.7}} \\			
			
			\hline
			
			\multicolumn{2}{c|}{\makecell[l]{VCTree \cite{VCTree}}} & \makecell[c]{65.4 / 67.2} & \makecell[c]{16.7 / 18.2} & \makecell[c]{26.6 / 28.6} & \makecell[c]{46.7 / 47.6}&\makecell[c]{11.8 / 12.5}&\makecell[c]{18.8 / 19.8} &\makecell[c]{31.9 / 36.2} &\makecell[c]{7.4 / 8.7} &\makecell[c]{12.0 / 14.0} \\

			\multicolumn{2}{c|}{\makecell[l]{  + TDE \cite{TDE}}} & \makecell[c]{47.2 / 51.6} & \makecell[c]{25.4 / 28.7} & \makecell[c]{33.0 / 36.9} & \makecell[c]{25.4 / 27.9}&\makecell[c]{12.2 / 14.0}&\makecell[c]{16.5 / 18.6} &\makecell[c]{19.4 / 23.2} &\makecell[c]{9.3 / 11.1} &\makecell[c]{12.6 / 15.1} \\

			\multicolumn{2}{c|}{\makecell[l]{  + EIL}} & \makecell[c]{64.5 / 66.5} & \makecell[c]{22.8 / 24.3} & \makecell[c]{33.7/ 35.6} & \makecell[c]{45.9 / 46.9}&\makecell[c]{\textbf{17.8 / 18.9}}&\makecell[c]{\textbf{25.6 / 26.9}} &\makecell[c]{31.2 / 35.5} &\makecell[c]{\textbf{10.6 / 12.4}} &\makecell[c]{\textbf{15.9 / 18.3}} \\

			\hline

			\multicolumn{2}{c|}{\makecell[l]{Transformer \cite{TDE}}} & \makecell[c]{63.6 / 65.7} & \makecell[c]{19.7 / 19.6} & \makecell[c]{27.9 / 30.2} & \makecell[c]{38.1 / 39.2}&\makecell[c]{9.9 / 10.5}&\makecell[c]{15.7 / 16.6} &\makecell[c]{30.0 / 34.3} &\makecell[c]{7.4 / 8.8} &\makecell[c]{11.9 / 14.0} \\	
			
			\multicolumn{2}{c|}{\makecell[l]{  + CogTree \cite{cogtree}}} & \makecell[c]{38.4 / 39.7} & \makecell[c]{28.4 / 31.0} & \makecell[c]{32.7 / 34.8} & \makecell[c]{22.9 / 23.4}&\makecell[c]{15.7 / 16.7}&\makecell[c]{18.6 / 19.5} &\makecell[c]{19.5 / 21.7} &\makecell[c]{11.1 / 12.7} &\makecell[c]{14.1 / 16.0} \\

			\multicolumn{2}{c|}{\makecell[l]{  + EIL}} & \makecell[c]{63.3 / 65.0} & \makecell[c]{27.7 / 29.8} & \makecell[c]{\textbf{38.6/ 40.9}} & \makecell[c]{38.2 / 40.0}&\makecell[c]{\textbf{15.7} / 16.5}&\makecell[c]{\textbf{22.3 / 23.2}} &\makecell[c]{31.7 / 36.1} &\makecell[c]{\textbf{12.7 / 14.9}} &\makecell[c]{\textbf{18.1 / 21.0}} \\			
			
			\hline

			\\
			
	\end{tabular}}
	
	\caption{Ablation study of the Environment-Invariant Learning  (EIL) module on VG dataset.}
	\label{ta:3}
\end{table*}

\textbf{Dataset}. In the SGG task, we choose Visual Genome (VG) \cite{VG} dataset which comprises 75k object categories and 40k predicate categories. We applied the widely accepted benchmark \cite{motifs,VCTree,TDE,hwzhang2017,icarl}, using the 150 highest frequency objects categories and 50 predicate categories. GQA \cite{gqa} is another  dataset for vision-language tasks with more than 3.8M relation annotations. Following previous work \cite{SHA}, we select Top-200 object classes as well as Top-100 predicate classes by their frequency for the GQA200 benchmark. For both datasets, the training set is set to be 70\%, and the testing set is 30\%, with 5k images from the training set for validation  \cite{motifs}.

\begin{table}
	\centering
	\scriptsize 
	\setlength{\tabcolsep}{1mm}{
		\begin{tabular}{ccc|ccc}
			\hline
			\multicolumn{3}{c|}{\textbf{Environments}}  & \multicolumn{3}{c}{\makecell[c]{\textbf{SGCls}}} \\
			
			\Xcline{4-6}{0.5pt}
			Normal & Balanced & Over-Balanced  &\makecell[c]R@50 / 100 &\makecell[c]mR@50 / 100& \makecell[c]F@50 / 100  \\

			\hline
			
			& & & \makecell[c]{38.9 / 39.8}&\makecell[c]{8.3 / 8.8}&\makecell[c]{13.7 / 14.8}  \\
			
			& \checkmark &\checkmark & \makecell[c]{21.5 / 22.6}&\makecell[c]{22.1 / 23.4}&\makecell[c]{21.8 / 23.0}  \\
			
			\checkmark&  &\checkmark & \makecell[c]{39.8 / 40.6}&\makecell[c]{13.7 / 14.8}&\makecell[c]{20.3 / 21.6}  \\
			
			\checkmark& \checkmark & & \makecell[c]{39.7 / 40.5}&\makecell[c]{13.3 / 14.0}&\makecell[c]{19.9 / 20.8}  \\			
			\hline
			
			\checkmark& \checkmark & \checkmark &\makecell[c]{39.3 / 40.1}& \makecell[c]{15.4 / 16.1}&\makecell[c]{\textbf{22.1 / 23.0}}\\

			\hline

			\\
			
	\end{tabular}}
	
	\caption{Ablation study of constructing different environments on VG dataset.}
	\label{ta:4}
\end{table}

\begin{table}
	\centering
	\footnotesize
	\setlength{\tabcolsep}{1mm}{
		\begin{tabular}{cc|ccc}
			\hline
			\multicolumn{2}{c|}{\multirow{2}*{\textbf{Model}}}  & \multicolumn{3}{c}{\makecell[c]{\textbf{SGCls}}} \\
			
			\Xcline{3-5}{0.5pt}
			&   &\makecell[c]R@50 / 100 &\makecell[c]mR@50 / 100& \makecell[c]F@50 / 100  \\

			\hline
			
			\multicolumn{2}{c|}{w/o-Curriculum Schedule} & \makecell[c]{39.3 / 40.1}& \makecell[c]{15.4 / 16.1}&\makecell[c]{22.1 / 23.0}  \\
			
			\multicolumn{2}{c|}{w/o-Norm Schedule}& \makecell[c]{39.2 / 40.0}&\makecell[c]{14.8 / 15.8}&\makecell[c]{21.5 / 22.7}  \\
			
			\multicolumn{2}{c|}{w/o-Over Schedule} & \makecell[c]{35.6 / 36.4}&\makecell[c]{18.1 / 19.1}&\makecell[c]{24.0 / 25.1}  \\
			\hline
			\multicolumn{2}{c|}{w-Curriculum Schedule} & \makecell[c]{34.5 / 35.4}&\makecell[c]{\textbf{20.8 / 21.8}}&\makecell[c]{\textbf{25.9 / 27.0}}  \\

			\hline

			\\
			
	\end{tabular}}
	
	\caption{Ablation study for curriculum learning strategy on VG dataset.}
	\label{ta:5}
\end{table}

\begin{table}
	\centering
	\small
	\setlength{\tabcolsep}{2mm}{
		\vspace{5pt}		
		\begin{tabular}{lc|ccc}
			\hline
			\multicolumn{2}{c|}{\multirow{2}*{\textbf{Setting}}}  & \multicolumn{3}{c}{\makecell[c]{\textbf{PredCls}}} \\
			
			\Xcline{3-5}{0.5pt}
			&   &\makecell[c]R@50 / 100 &\makecell[c]mR@50 / 100& \makecell[c]F@50 / 100  \\

			\hline

			\multicolumn{2}{l|}{w/o IRM term}& \makecell[c]{52.4 / 54.4}&\makecell[c]{35.4 / 37.4}&\makecell[c]{42.2 / 44.3}  \\
			
			\multicolumn{2}{l|}{w/ IRM term} & \makecell[c]{\textbf{55.3 / 57.4}}&\makecell[c]{34.9 / 37.0}&\makecell[c]{\textbf{42.8 / 45.0}}  \\

			\hline
			
			\\

	\end{tabular}}
	\vspace{-5pt}
	\caption{Ablation study of the IRM term.}
	\label{ta:7}
\end{table}

\textbf{Tasks}. Following previous works \cite{motifs,VCTree,SHA,recovering}, we evaluate our model on three widely used SGG tasks: (1) Predicate Classification (PredCls): given images, object bounding boxes, and object labels, models are required to recognize predicate classes. (2) Scene Graph Classification (SGCls): gives images and object bounding boxes and asks models to predict object labels and relationship labels between objects. (3) Scene Graph Detection (SGDet): models are required to localize objects, recognize objects, and predict their relationships directly from images.

\textbf{Metrics}. Following previous works \cite{BGNN,PCPL}, we use Recall@K (R@K) and mean Recall@K (mR@K) as our metrics. Moreover, inspired by previous work \cite{IETrans}, we use the overall metric F@K  to jointly evaluate R@K and mR@K, which is the harmonic average of R@K and mR@K.

\textbf{Implementation Details}. We employ a pre-trained Faster-RCNN \cite{fasterRCNN} with ResNeXt-101-FPN \cite{FPN} provided by \cite{TDE} as the object detector. We use Glove \cite{Glove} to obtain the semantic embedding. In the training process, the parameters of the detector are fixed to reduce the computation cost. The hyper-parameter lambda which balances the different environments is set to 0.9. We optimize all models with an Adam optimizer with a momentum of 0.9. The batch size is set to 4, and the total training stage lasts for 120,000 steps with $T=30000$ and $\lambda_{max}=0.9$. The initial learning rate is 0.001, and we adopt the same warm-up and decayed strategy as \cite{SHA}. One RTX2080 Ti is used to conduct all the experiments.

\subsection{Compared Methods}

We demonstrate the effectiveness of our method by comparing the results with current SOTA methods and validate its generalizability with different baseline models. On the one hand, to prove its performance, we select some dedicated designed SGG models with state-of-the-art performance, including re-produced IMP \cite{imp}, GPS-Net \cite{GPSNet}, DT2-ACBS \cite{DT2-ACBS}, SHA-GCL \cite{SHA}, and BGNN \cite{BGNN}. On the other hand, to demonstrate the generalizability of our EICR, we compare our method with the model-agnostic baselines which can be applied in a plug-and-play fashion, including TDE \cite{TDE}, CogTree \cite{cogtree}, PCPL \cite{PCPL}, EBM \cite{EBM}, DLFE \cite{DLFE}, GCL \cite{SHA}, NICE \cite{NICE} and IETrans \cite{IETrans}.

\begin{table}
	\centering
	\small
	\setlength{\tabcolsep}{2mm}{
		\begin{tabular}{cc|ccc}
			\hline
			\multicolumn{2}{c|}{\multirow{2}*{\textbf{$\lambda_{max}$}}}  & \multicolumn{3}{c}{\makecell[c]{\textbf{SGCls}}} \\
			
			\Xcline{3-5}{0.5pt}
			&   &\makecell[c]R@50 / 100 &\makecell[c]mR@50 / 100& \makecell[c]F@50 / 100  \\

			\hline
			
			\multicolumn{2}{c|}{w/o-$\lambda_{max}$} & \makecell[c]{39.3 / 40.1}& \makecell[c]{15.4 / 16.1}&\makecell[c]{22.1 / 23.0}  \\
			\hline
			\multicolumn{2}{c|}{0.7}& \makecell[c]{37.1 / 38.0}&\makecell[c]{18.0 / 19.0}&\makecell[c]{24.3 / 25.2}  \\
			
			\multicolumn{2}{c|}{0.8} & \makecell[c]{36.2 / 37.1}&\makecell[c]{19.1 / 20.0}&\makecell[c]{25.0 / 26.0}  \\
			
			\multicolumn{2}{c|}{0.87} & \makecell[c]{34.8 / 35.6}&\makecell[c]{18.9 / 19.7}&\makecell[c]{24.5 / 25.4}  \\		
			
			\multicolumn{2}{c|}{0.9} & \makecell[c]{34.5 / 35.4}&\makecell[c]{20.8 / 21.8}&\makecell[c]{25.9 / 27.0}  \\	
			
			\multicolumn{2}{c|}{0.92} & \makecell[c]{32.9 / 33.8}&\makecell[c]{21.1 / 22.1}&\makecell[c]{25.7 / 26.7}  \\		
			
			\multicolumn{2}{c|}{0.95} & \makecell[c]{33.3 / 34.2}&\makecell[c]{20.0 / 21.1}&\makecell[c]{25.0 / 26.1}  \\	
			
			\multicolumn{2}{c|}{0.99} & \makecell[c]{27.6 / 28.7}&\makecell[c]{20.9 / 21.9}&\makecell[c]{23.8 / 24.8}  \\	
			
			\hline

			\\
			
	\end{tabular}}
	
	\caption{Parameter analysis towards $\lambda_{max}$ on VG dataset.}
	\label{ta:6}
\end{table}

\begin{figure*}[t]
	\begin{center}
		\includegraphics[width=1.0\linewidth]{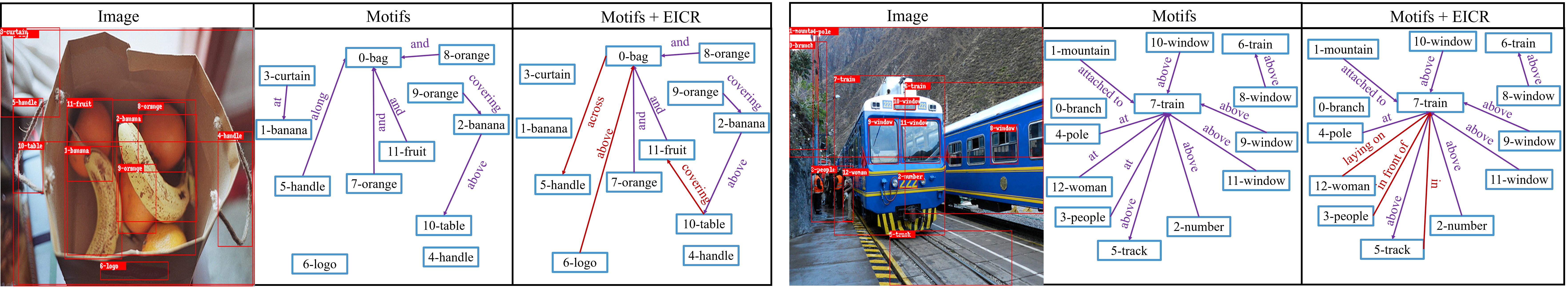}
	\end{center}
	\caption{Visualization scene graphs between Motifs and Motifs + EICR with regard to R@20 on PredCls setting. Purple edges represent the reasonable relationships predicted by Motifs. Red edges represent  the refined reasonable relationships which are predicted by Motifs + EICR but failed to be detected by Motifs.}
	\label{fig:fig5}
\end{figure*}
\begin{figure*}[t]
	\begin{center}
		\includegraphics[width=1.0\linewidth]{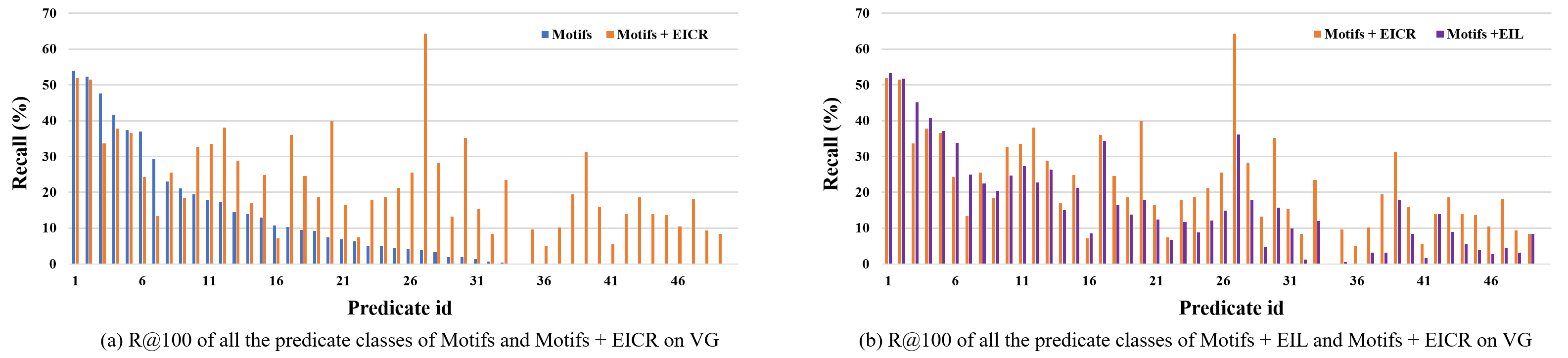}
	\end{center}
	\caption{R@100 of 50 predicate classes on SGCls on the VG dataset.}
	\label{fig:fig4}
\end{figure*}

\subsection{Main Results}

We report the results of the proposed EICR framework and baselines on the VG and GQA datasets in Table ~\ref{ta:1} and Table ~\ref{ta:2}. From the results of various tasks and baselines, we have several observations as follows:  

On the one hand, our EICR  is adaptive to different baselines. We adapt our method to 3 popular baselines for SGG, including Motifs \cite{motifs}, VCTree \cite{VCTree}, and Transformer \cite{TDE}. These baselines include various architectures such as conventional LSTM (Motifs), tree structure (VCTree), and self-attention layers (Transformer). Various training algorithms are also contained such as supervised training and reinforcement learning (VCTree). Specifically, our method can boost all models’ mR@50/100 metric and the overall F@50/100 metric. With our method, the results for VCTree are improved over 14\% on mR@50/100 and improved over 12\% across all 3 tasks on the metric F@50/100. Moreover, compared with other model-agnostic methods, our method outperforms all of them on the F@50/100 and gains competitive results on mR@50/100. By applying our methods to VCTree and Transformer on SGCls and SGDet, our model can achieve the highest R@50/100 and mR@50/100 among all model-agnostic baselines.

On the other hand, compared with strong specific baselines, our method can also achieve competitive performance on mR@50/100 and the best overall performance on F@50/100. Our method with VCTree is close to the SOTA results in DT2-ACBS on SGCls and SGDet tasks on mR@50/100 while outperforming much better than them on R@50/100.  For an overall comparison of the F@50/100 metrics, our method with VCTree can achieve the best F@50/100 on PredCls and SGCls and our method with Motif achieves the best F@50/100 in the SGDet task.

\subsection{Ablation Studies}

In this part, we analyze the influence of the environment-invariant learning, curriculum learning strategy, and corresponding parameter $\lambda_{max}$.

\textbf{Influence of Environment-Invariant Learning (EIL)}. Table ~\ref{ta:3} and Table ~\ref{ta:4} present the results of all the ablation models.   As shown in Table ~\ref{ta:3},  only using environment-invariant learning is hard to boost the mR@50/100 and F@50/100 performance as much as EICR. The reason is that the training procedure is still led by the normal environments and overfitting the corresponding general patterns. However, though the performance on the mR@50/100 and F@50/100 are not boosted so much,  the EIL retains the performance on the R@50/100 compared with the original baselines (Motifs, VCTree, Transformer). We can conclude that by introducing EIL to cope with the context imbalance problem, the model learns the context-unbiased relation representation and make a gain on the mR@50/100 metric while retaining the general patterns without the decrease on the R@50/100 metric. As shown in Table ~\ref{ta:4},  different settings of learning environments all achieve improvements on mR@50/100 and F@50/100 compared with Motifs. However, its performance is poor compared with EIL, which shows the importance of combing multiple environments by EIL. Integration of multiple learning environments can alleviate the context imbalance and improve the performance on SGG benchmarks.

\textbf{Influence of Curriculum Learning Strategy}. As aforementioned, we propose the class-balanced curriculum learning strategy to alleviate the class imbalance. In order to prove the effectiveness of the above components, we test various ablation models on the VG dataset as follows:

(1) w/o-Curriculum Schedule: To evaluate the effectiveness of the curriculum schedule, we do not use curriculum schedule, i.e., $\mathcal{R}_{\text {hybird}}=  \mathcal{R}_{norm}+ \mathcal{R}_{over}+ \mathcal{R}_{balanced}$. 

(2) w/o-Norm Schedule: To evaluate the effectiveness of the changing weight of the normal environment, we remove the curriculum schedule for the normal environment risk and only employ the curriculum schedule for the over-balanced environment, i.e., $\mathcal{R}_{\text {hybird}}= \mathcal{R}_{norm}+(1-\lambda) \cdot \mathcal{R}_{over}+ \mathcal{R}_{balanced}$. 

(3) w/o-Over Schedule: To evaluate the effectiveness of the curriculum schedule for the over-balanced environment, we remove the curriculum schedule for the over-balanced environment, i.e., $\mathcal{R}_{\text {hybird}}=\lambda \cdot \mathcal{R}_{norm}+ \mathcal{R}_{over}+ \mathcal{R}_{balanced}$.

Table ~\ref{ta:5} presents the results of all the ablation models. First, the curriculum schedule can achieve a huge improvement on the mR@50/100 and  F@50/100 metrics. Compared with w/o-curriculum schedule, w-curriculum schedule boosts the mR@50/100 metric by over 5 points and improves by nearly 4 points on the F@50/100. Second, we witness an obvious performance decay when removing the curriculum schedule either for the normal environment or the over-balanced environment. It verifies that constructing curriculum learning schedules for multiple environments would effectively alleviate the class imbalance in the SGG dataset, thus leading to class-unbiased relation predictions.

\textbf{Influence of IRM regularization}. We take Motifs \cite{motifs} as the baseline model. As shown in Table ~\ref{ta:3}, we can see that adding the IRM regularization term improves the performance of R@50/100 and F@50/100, demonstrating that the IRM regularization enhances the representation ability of the predicate predictor.

\begin{figure}[t]
	\begin{center}
		\includegraphics[width=1.0\linewidth]{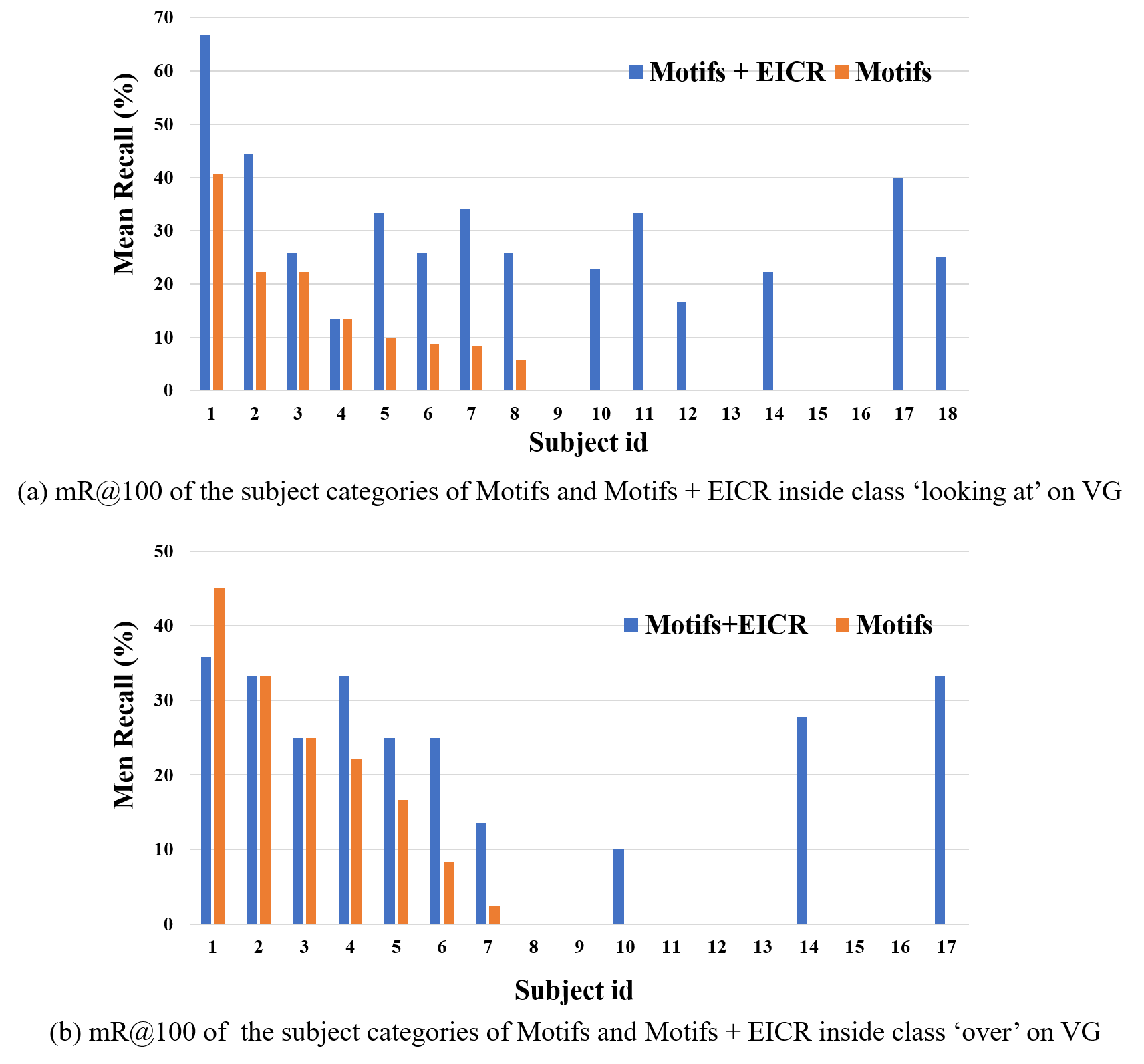}
	\end{center}
	\caption{mR@100 of various subject categories inside the predicate classes on the VG dataset.}
	\label{fig:fig6}
\end{figure}

\textbf{Influence of $\lambda_{max}$}. As shown in Table ~\ref{ta:6}, the mR@50/100 metric significantly increases with the increase of the $\lambda_{max}$ while the R@50/100 metric decreases at the same time.  Since the over-balanced environment is highly related to the samples from the tail predicates, the increase of the $\lambda_{max}$ can somewhat be considered as increasing the model's attention to the tail samples.  Thus, the phenomenon indicates that the conventional structures of the SGG model (Motifs, VCTree, Transformer) may easily classify tail classes as negative samples and lead to low results on mR@50/100, while this part of the data is of vital significance for improving the model’s ability to make class-unbiased predictions.

\begin{table}
	\centering
	\small
	\setlength{\tabcolsep}{2mm}{
		\begin{tabular}{lc|ccc}
			\hline
			\multicolumn{2}{c|}{\multirow{2}*{\textbf{Method}}}  & \multicolumn{3}{c}{\makecell[c]{\textbf{PredCls}}} \\
			
			\Xcline{3-5}{0.5pt}
			&   &\makecell[c]R@50 / 100 &\makecell[c]mR@50 / 100& \makecell[c]F@50 / 100  \\
			
			\hline
			
			\multicolumn{2}{c|}{BBN \cite{BBN}} & \makecell[c]{56.0 / 57.7}& \makecell[c]{19.4 / 21.3}&\makecell[c]{28.8 / 31.1}  \\
			
			\multicolumn{2}{c|}{Reweight \cite{reweight}}& \makecell[c]{54.7 / 56.5}&\makecell[c]{17.3 / 18.6}&\makecell[c]{26.3 / 28.0}  \\
			
			\multicolumn{2}{c|}{EICR} & \makecell[c]{55.3 / 57.4}&\makecell[c]{\textbf{34.9 / 37.0}}&\makecell[c]{\textbf{42.8 / 45.0}}  \\

			\hline
			\\

	\end{tabular}}
	\vspace{-5pt}
	\caption{Related class-balancing strategies on VG dataset.}
	\vspace{-4pt}
	\label{ta:8}
\end{table}

\begin{table}
	\centering
	\small
	\setlength{\tabcolsep}{1.5mm}{
		\begin{tabular}{cc|ccc}
			\hline
			\multicolumn{2}{c|}{\multirow{2}*{Model}}  & \makecell[c]{PredCls}& \makecell[c]{SGCls}& \makecell[c]{SGDet} \\
			
			\Xcline{3-5}{0.5pt}
			&   &\makecell[c]mT@50 / 100 &\makecell[c]mT@50 / 100& \makecell[c]mT@50 / 100  \\

			\hline
			
			\multicolumn{2}{c|}{Motifs} & \makecell[c]{7.9 / 8.8}& \makecell[c]{3.1 / 3.4}&\makecell[c]{2.0 / 2.4}  \\
			
			\multicolumn{2}{c|}{ + EICR} & \makecell[c]{17.8 / 19.2}& \makecell[c]{8.3 / 8.9}&\makecell[c]{5.8 / 6.6}  \\

			\hline

			\multicolumn{2}{c|}{VCTree} & \makecell[c]{8.4 / 9.3}& \makecell[c]{4.3 / 4.8}&\makecell[c]{1.7 / 2.1}  \\
			
			\multicolumn{2}{c|}{ + EICR} & \makecell[c]{18.3 / 19.7}& \makecell[c]{11.6 / 12.4}&\makecell[c]{5.8 / 6.7}  \\

			\hline
			
			\multicolumn{2}{c|}{Transformer} & \makecell[c]{9.6 / 10.6}& \makecell[c]{3.3 / 3.7}&\makecell[c]{2.4 / 2.9}  \\
			
			\multicolumn{2}{c|}{ + EICR} & \makecell[c]{18.8 / 20.3}& \makecell[c]{8.9 / 9.4}&\makecell[c]{6.7 / 7.6}  \\

			\hline

			\\
			
	\end{tabular}}
	
	\caption{Performance of balancing the contexts of our EICR method on VG dataset.}
	\label{ta:s1}
\end{table}

\subsection{Qualitative Studies}

To get an intuitive perception of the superior performance on the SGG tasks of our proposed method, we make quantitative studies.

\textbf{Visualization}. To show the potential of our method for real-world application, we visualize several PredCls examples generated from the biased Motifs and the unbiased Motifs + EICR. As shown in Fig.~\ref{fig:fig5}, we can observe that our method can help to generate more various relation predicates while keeping faithful to the image content. The model prefers to provide more specific relationship predictions (e.g., `covering' and `in front of') rather than common and trivial ones (e.g., `at' and `along').  Moreover, our method could also help capture potential reasonable relationships. For example,  in Fig.~\ref{fig:fig5}, our method captures  `logo-on-bag' in the left example and `track-in-train' in the right example. In a nutshell, the proposed method could enhance the unbiased scene graph generation and generate more informative relation triplets to support various downstream tasks.

\textbf{Detailed Results}. In Fig.~\ref{fig:fig4} (a), we show the detailed results of comparing Motifs and Motifs + EICR with respect to R@100 of all the predicate classes on the SGCls task. We can observe that by applying our methods to the Motifs baseline, though there exists an acceptable decay on the minority of several head predicate classes, the performance on most of the predicate classes is obviously improved.  Moreover, we also compare Motifs + EIL (i.e., w/o curriculum schedule) and Motifs + EICR on the detailed performance towards every predicate class on VG. As shown in Fig.~\ref{fig:fig4} (b), the curriculum schedule effectively prevents the model from overfitting the general pattern on the head predicate classes and achieves a better performance towards the tail predicate predictions. It demonstrates that the curriculum schedule could achieve a reasonable trade-off between the environments, and effectively alleviate the class imbalance of the predicates to get a class-unbiased relation classifier.

\subsection{Further Analysis}

\begin{figure}[t]
	\begin{center}
		\includegraphics[width=0.9\linewidth]{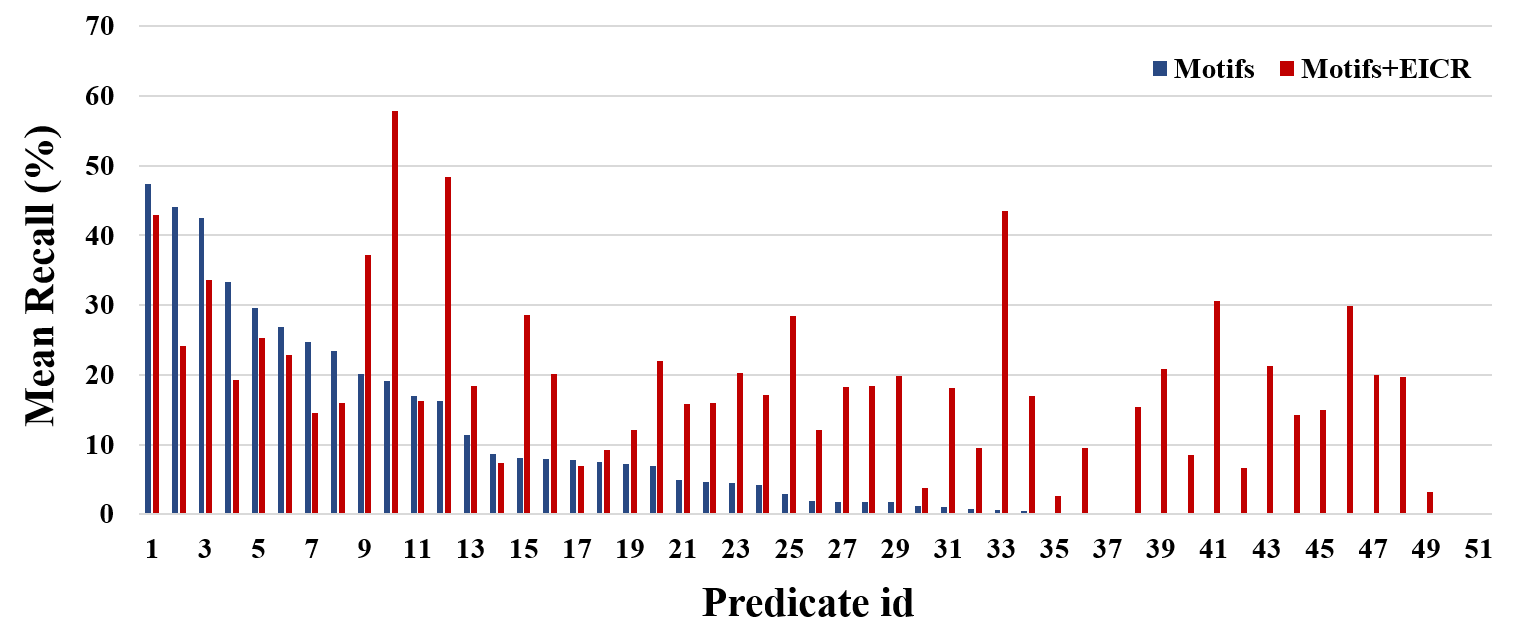}
	\end{center}
	\caption{mR@100 of various triplets with different contexts inside all predicate classes of Motifs and Motifs + EICR on VG.}
	\label{fig:figs2}
\end{figure}

\textbf{Verification for Alleviating the Context Imbalance}.  To make a further analysis, we verify the effectiveness of our  EICR for alleviating the context imbalance between the various subject-object pairs. Specifically, we calculate the  mR@100 of all the different subject categories inside the same predicate class. i.e., the mR@100 for the subject `man' is calculated by the mean R@100 of the relation triplets with the subject `man' such as `man-wearing-shirt' and `man-wearing-boots'.  Two examples on the VG dataset are shown in Fig.~\ref{fig:fig6}. We can see that compared with the Motifs,  the EICR help to make a more balanced distribution for the various subject-object pairs thus gaining context-unbiased results.

\textbf{Discussions with Relevant Long-Tailed Approaches}. To demonstrate the effectiveness of our EICR strategy in alleviating the class imbalance, we compare our method with two other relevant typical class-balancing strategies on the Motifs baseline, i.e., the resampling strategy BBN \cite{BBN}  and the reweighting strategy \cite{reweight} following SHA \cite{SHA}. As shown in Table \ref{ta:8}, we can see that our method achieves the best performance. We can see that compared with the Motifs,  our EICR model could help to make a more balanced distribution for the various predicate classes thus gaining better results on the various SGG datasets.

\textbf{Verification for Balancing Contexts}. To provide a more detailed analysis of our method's effectiveness in alleviating the context imbalance, we report the metric mT@50/100 on the VG dataset. mT@50/100 denotes the average of the mean Recall for various triplets (i.e., the same predicate with different subject-object context) inside each predicate class. As shown in Table ~\ref{ta:s1}, our EICR can be applied in a plug-and-play fashion for solving the context imbalance. By adding our EICR to the three baselines,  the results are significantly improved across all 3 tasks on the metric mT@50/100. Moreover, in Fig.~\ref{fig:figs2}, we show the detailed results of comparing Motifs and Motifs + EICR with respect to mR@100 of the triplets inside all predicate classes on the PredCls task. With our method,  the performance of the triplets inside most of the predicate classes is obviously improved. It demonstrates that our method could achieve a reasonable trade-off for the existing imbalance contexts between the predicate classes, and effectively alleviate the context imbalance.

\textbf{Influence of $T$}. To provide a more detailed analysis of the influence of the curriculum learning module, we report the performance with different $T$. As shown in Table ~\ref{ta:s2}, with the increase of the intermediate training iterations $T$, the mR@50/100 metric and the overall metric F@50/100 first increases and then decreases. The phenomenon shows that blindly focusing on the tail predicates does not necessarily mean higher performance on the various SGG datasets.

\begin{table}
	\centering
	\small
	\setlength{\tabcolsep}{2mm}{
		\begin{tabular}{cc|ccc}
			\hline
			\multicolumn{2}{c|}{\multirow{2}*{\textbf{$T$}}}  & \multicolumn{3}{c}{\makecell[c]{\textbf{SGCls}}} \\
			
			\Xcline{3-5}{0.5pt}
			&   &\makecell[c]R@50 / 100 &\makecell[c]mR@50 / 100& \makecell[c]F@50 / 100  \\

			\hline
			
			\multicolumn{2}{c|}{10000}& \makecell[c]{34.3 / 35.1}&\makecell[c]{19.9 / 20.9}&\makecell[c]{25.2 / 26.2}  \\
			
			\multicolumn{2}{c|}{20000} & \makecell[c]{34.3 / 35.1}&\makecell[c]{20.8 / 21.6}&\makecell[c]{25.9 / 26.8}  \\
			
			\multicolumn{2}{c|}{30000} &  \makecell[c]{34.5 / 35.4}&\makecell[c]{20.8 / 21.8}&\makecell[c]{25.9 / 27.0}   \\		
			
			\multicolumn{2}{c|}{40000} & \makecell[c]{34.9 / 35.8}&\makecell[c]{19.0 / 19.9}&\makecell[c]{24.6 / 25.6}  \\

			\hline

			\\
			
	\end{tabular}}
	
	\caption{Parameter analysis towards $T$ on VG dataset.}
	\label{ta:s2}
\end{table}

\section{Conclusions} 
In this paper, we design a method named EICR for fine-grained scene graph generation. We were motivated by the observation that there not only exists the class imbalance between predicate classes, but also the context imbalance for various subject-object pairs.   The proposed EICR consists of two debias modules to learn a robust relation classifier unbiased to the various class and contexts. Comprehensive experiments show the effectiveness of our method.  In the future, we will further analyze more effective methods to alleviate the context imbalance and explore our theory in other visual recognition problems (e.g., image classification) with similar challenges.

\textbf{Acknowledgement}. Our work was supported by Joint Fund of Ministry of Education of China (8091B022149). Key Research and Development Program of Shanxi (2021ZDLGY01-03). National Natural Science Foundation of China (62102293, 6213201662171343, and 62071361) and Fundamental Research Funds for the Central Universities (ZDRC2102).

{\small
	\bibliographystyle{ieee_fullname}
	\bibliography{egbib}
}

\end{document}